\def\year{2020}\relax
\documentclass[letterpaper]{article} 
\usepackage{aaai20}  
\usepackage{times}  
\usepackage{helvet} 
\usepackage{courier}  
\usepackage[hyphens]{url}  
\usepackage{graphicx} 
\usepackage{graphicx}  
\usepackage{enumitem}
\usepackage{amsmath}
\usepackage{amsthm}
\usepackage{amsfonts} 
\usepackage{amssymb}
\usepackage{mathrsfs}
\usepackage{fixmath}

\usepackage{tabularx}
\usepackage{threeparttable}
\usepackage{setspace}
\usepackage{multirow}

\usepackage{appendix}

\newtheorem{theorem}{Theorem}

\newtheorem{prop}{Proposition}

\usepackage{caption}
\newcommand{\xhdr}[1]{{\noindent\bfseries #1}.}
\title{Improving Graph Attention Networks with\\ Large Margin-based Constraints}
\author{
  Guangtao Wang$^{1}$, Rex Ying$^{2}$, Jing Huang$^{1}$, Jure Leskovec$^{2}$\\
\Large \textbf{$^{1}$JD AI Research, Mountain View, CA}\\
\Large \textbf{$^{2}$Department of Computer Science, Stanford University, Stanford, CA}\\
$^{1}$\texttt{\{guangtao.wang, jing.huang\}@jd.com}\\ $^{2}$\{\texttt{rexying@stanford.edu, jure@cs.stanford.edu}
}
 
\begin{document}

\maketitle

\begin{abstract}

Graph Attention Networks (GATs) are the state-of-the-art neural architecture for representation learning with graphs. GATs learn attention functions that assign weights to nodes so that different nodes have different influences in the feature aggregation steps. In practice, however, induced attention functions are prone to over-fitting due to increasing number of parameters and the lack of direct supervision on attention weights. GATs also suffer from over-smoothing at the decision boundary of nodes. Here we propose a framework to address their weaknesses via margin-based constraints on attention during training. We first theoretically demonstrate the over-smoothing behavior of GATs and then develop an approach using constraint on the attention weights according to the class boundary and feature aggregation pattern.
Furthermore, to alleviate the over-fitting problem, we propose additional constraints on graph structure. Extensive experiments and ablation studies on common benchmark datasets demonstrate the effectiveness of our method, which leads to significant improvements over the previous state-of-the-art graph attention methods on all datasets.
\end{abstract}

\section{Introduction}
Many real world applications involve graph data, like social networks~\cite{zhang2018link}, chemical molecules~\cite{gilmer2017neural}, and recommender systems~\cite{berg2017graph}. The complicated structures of these graphs have inspired new machine learning methods~\cite{cai2018comprehensive,wu2019comprehensive}. Recently much attention and progress has been made on graph neural networks, which have been successfully applied to social network analysis~\cite{battaglia2016interaction}, recommendation systems~\cite{ying2018graph}, and machine reading comprehension~\cite{tu2019Multi,de2018question}.

Recently, a novel architecture leveraging attention mechanism in Graph Neural Networks (GNNs) called Graph Attention Networks (GATs) was introduced~\cite{velivckovic2017graph}. GAT was motivated by attention mechanism in natural language processing~\cite{vaswani2017attention,devlin2018bert}. It computes representation of each node by attending to its neighbors via a masked self-attention. For each node, different weights are learned by attention functions so that the nodes in the same neighborhood have different weights in the feature aggregation step. Inspired by such attention-based architecture, several new attention-based GNNs have been proposed, and have achieved state-of-the-art performance on node classification benchmarks~\cite{liu2018geniepath,zhang2018gaan,ryu2018deeply,lee2018graph,thekumparampil2018attention,abu2018watch}.

However, attention-based GNNs suffer from the problems of overfitting and over-smoothing: (1) learned attention functions that use node features to assign an importance weight to every neighboring node tend to overfit the training data because the masked self-attention forces attention weights to be only computed for direct neighbors. (2) The over-smoothing problem arises for nodes that are connected but lie on different sides of the class decision boundary. Due to information exchanging over these edges, stacking multiple attention layers causes excessive smoothing of node features, and makes nodes from different classes become indistinguishable.

Here we develop a framework called {\em Constrained Graph Attention Networks (C-GATs)} that address the above shortcomings of GATs via the use of margin-based constraints. Margin-based constraints act as regularizers that prevent the over-fitting of GATs. For example, by adding the constraint that the learned attention weights of the nodes in the neighborhood should be greater than those of the nodes in non-neighborhood by a large margin, we guide the attention weights to separate one node's neighboring nodes and non-neighboring nodes by a margin, while attention weights in GATs is not capable of doing the same. This helps the model to learn the attention function which generalizes well to unseen graph structures.

To overcome the problem of over-smoothing, we propose constraints on attentions based on class labels, and a new feature aggregation function which only selects the neighbors with top $k$ attention weights for feature aggregation. The purpose of the proposed aggregation function is to reduce information propagation between nodes belonging to different classes.

In order to train the proposed C-GAT model efficiently and effectively, we develop a layer-wise adaptive negative sampling strategy. In contrast to the uniform negative sampling that suffers from the problem of inefficiency due to the fact that many negative samples do not provide any meaningful information, our negative sampling method obtains highly informative negative nodes in a layer-wise adaptive way.

We evaluate the proposed approach on four node-classification benchmarks: Cora, Citeseer, and Pubmed citation networks as well as an inductive protein-protein interaction (PPI) dataset. Extensive experiments demonstrate the effectiveness of our approach regarding to the classification accuracy and generalization on unseen graph structure: our C-GAT models improve consistently over the state-of-the-art GATs on all four datasets, especially with a new state-of-the-art accuracy number $98.8\%$ on the inductive learning PPI data. 

In summary, we make the following contributions in this paper:
\begin{itemize} [leftmargin=5pt,noitemsep]
	\item We provide new insights and mathematical analysis of the attention based GNN models for node classification and associated challenges;
	\item We propose a new constrained graph attention network (C-GAT) that utilizes constraints over the node attentions to overcome the problem of over-fitting and over-smoothing. And we propose an adaptive layer-wise negative sampling strategy to train C-GAT efficiently and effectively; 
	\item We propose an aggregation strategy to further remedy the over-smoothing problem at the class boundary by selecting top $k$ neighbors for feature aggregation;
	\item Our extensive experimental results and analysis demonstrate the benefit of the C-GAT model and show consistent gains over state-of-the-art graph attention models on standard benchmark datasets for graph node classification. 
\end{itemize}

\section{Related Works}
GNNs can be generally divided into two groups: spectral and non-spectral models~\cite{cai2018comprehensive,hamilton2017inductive,velivckovic2017graph}, according to the type of convolution operations on graphs. The former generates convolution operations based on Laplacian eigenvectors~\cite{bruna2013spectral,henaff2015deep,defferrard2016convolutional}, and these models are usually difficult to generalize to graph with unseen structures~\cite{monti2017geometric}. The non-spectral methods generate convolution operations directly based on spatially close neighbors and usually exhibit better performance on unseen graphs~\cite{duvenaud2015convolutional,atwood2016diffusion,hamilton2017inductive,niepert2016learning,monti2017geometric}. Our algorithm conceptually belongs to the non-spectral approaches.

\noindent\textbf{Graph Convolutional Neural Networks (GCNs)} generalize  convolution operations from traditional image data to graphs. The key point is to find a function generating node's representation by aggregating its own features as well as neighbors' features~\cite{wu2019comprehensive}. Example models include SSE~\cite{dai2018learning} MPNN~\cite{gilmer2017neural}, GraphSAGE~\cite{hamilton2017inductive}, DCNN~\cite{atwood2016diffusion}, StoGCN~\cite{chen2017stochastic}, LGCN~\cite{gao2018large}, and more. These GCNs usually treat all nodes of the same neighborhood equally for the purpose of feature aggregation. 

\noindent\textbf{Graph Attention Networks (GATs)} generalize attention operation to graph data. GATs allow for assigning different importance to nodes of a same neighborhood at the feature aggregation step and increase the model capacity of GNNs~\cite{velivckovic2017graph}. Based on such framework, different attention-based GNN architectures have been proposed. Examples include GaAN~\cite{zhang2018gaan}, AGNN~\cite{thekumparampil2018attention}, GeniePath~\cite{liu2018geniepath}, and others. Different models usually use different attention functions to compute the importance of the nodes in neighborhood. However, such attention functions suffer from over-fitting problem in learning the attention weights. If there are edges between different clusters, these GNNs easily lead to over-smoothing of node's representation, which hurts the performance on downstream node classification task.

\section{Analysis of GATs}\label{sec:gatAnalysis}
In this section we briefly review the GAT model and identify its weaknesses.

\xhdr{Notation} Let $\mathcal{G} = (\mathcal{V}, \mathcal{E}, \mathbold{X})$ be a graph where $\mathcal{V}$ is the set of $N$ nodes (or vertices), $\mathcal{E}  \subseteq \mathcal{V} \times \mathcal{V}$ is the set of $M$ edges connecting $M$ pairs of nodes in $\mathcal{V}$, and $\mathbold{X} \in \mathbb{R}^{N\times d}$ represents the node input features, where each row $\mathbold{x}_{i}$ = $\mathbold{X}_{i:}$ is a $d$-dimensional vector of attribute values of node $v_i \in \mathcal{V}$ ($1\leq i \leq N$). In this paper, we consider undirected graphs. Suppose $\mathbold{A}_{N\times N}$ is the adjacency/weighted adjacency matrix of $\mathcal{G}$ with $\mathbold{A}_{i,j} \geq 0$, $\mathbold{D} = \text{diag}(d_1, d_2, \cdots, d_{N})$ and $d_i = \sum_{j=1}^{N}\mathbold{A}_{i,j}$. The graph Laplacian of $\mathcal{G}$ is defined as $\mathbold{L} = \mathbold{D} - \mathbold{A}$. And the random walk normalized Laplacian $\mathbold{L}_{rw} = \mathbold{D}^{-1}\mathbold{L}$.

\xhdr{Node classification} Suppose that $\mathcal{V}_{l} \subset \mathcal{V}$ consists of a set of labeled nodes, the goal of node classification is to predict the labels of the remaining unlabeled nodes. Many graph-based node labeling methods make the cluster assumption which assumes the connected nodes in the graph are likely to share the same label~\cite{weston2012deep,li2018deeper}.

\xhdr{Attention based GNN} utilizes the following layer-wise attention based aggregate function for node embedding on each node $v_i \in \mathcal{V}$:
\begin{flalign}
\mathbold{h}_{i}^{(l+1)} = \sigma(\sum_{j \in  \mathcal{N}_{i}}\alpha_{i,j}^{(l)}\mathbold{W}^{(l)}\mathbold{h}_{j}^{(l)})\nonumber\\
\alpha_{i,j} ^{(l)}= \frac{\exp(\phi_{\omega}^{(l)}(\mathbold{h}_{i}^{(l)}, \mathbold{h}_{j}^{(l)}))}{\sum_{k \in  \mathcal{N}_{i}}\text{exp}( \phi_{\omega}^{(l)}(\mathbold{h}_{i}^{(l)}, \mathbold{h}_{k}^{(l)}))}\label{eq:GAT}
\end{flalign}
Where $\mathbold{W}^{(l)} \in \mathbb{R}^{d^{(l+1)}\times d^{(l)}}$ is a  trainable weight matrix shared by $l$-th layer. $\sigma$ is the activation function. $\mathbold{h}_{i}^{(l)} \in \mathbb{R}^{d^{(l)}}$ is the node embedding achieved in $l$-th layer;  $\mathbold{h}_{i}^{(0)} = \mathbold{x}_{i}$. $ \mathcal{N}_{i}$ is the set of $v_i$'s one-hop neighboring nodes and also includes $v_i$ (i.e. there is a self-loop on each node). $\alpha^{(l)}_{i,j}$ is the $l$-th attention weight between the target node $v_i$ and the neighboring node $v_j$, which is generated by applying softmax to the values computed by attention function $\phi_{\omega}^{(l)}$, and $\omega$ is the trainable parameters of the attention function. For GAT~\cite{velivckovic2017graph} model, $\phi_{\omega}^{(l)} = \text{LeakyReLU}({\mathbold{a}}\cdot [\mathbold{W}^{(l)}\mathbold{h}_{i}^{(l)} \Vert \mathbold{W}^{(l)}\mathbold{h}_{j}^{(l)}])$, $\Vert$ is concatenation as in~\cite{velivckovic2017graph}. In this paper, we use GAT to refer to all attention-based GNN models.

\xhdr{The overfitting problem of GATs}
The attention functions in GAT compute the attention values based on the features of pairs of connected nodes (see Eq.~\ref{eq:GAT}). To train such attention functions in GAT, there is only one source to guide their parameters: the classification error. In other words, supervised information to learn these attention functions only comes from the labels of the nodes. 


There are two common sources of over-fitting in machine learning: 1) lack of enough supervision information for parameter learning, and 2) the model is over-parameterized. we believe that the over-fitting of GAT comes from the former source: lack of enough supervision data.
The supervision of GAT to learn attention parameters is limited and indirect, since the GAT supervision signal can only come from the $O(|V|)$ nodes labels for node classification. In general, smaller number of supervisions leads to more overfitting~\cite{trevor2009elements}.
The learned attention function performs well on the training data but fails to generalize and is not robust to perturbation. 
We demonstrate this in experimental section with robustness test.

\xhdr{The over-smoothing problem of GATs} To facilitate the analysis, we focus on the attention aggregation and simplify Eq. \ref{eq:GAT} in terms of matrix operation as $\mathbold{Y} = \mathbold{A} \mathbold{X}$\footnote{
Similar to \cite{li2018deeper}, we omit the non-linearity activation function $\sigma$. In fact \cite{wu2019simplifying} shows evidence that similar performance is observed in the case when there is no nonlinearity after the aggregation step.}, where $\mathbold{A}_{N\times N}$ is the attention matrix, $\mathbold{A}^{(l)}_{ij} =  \alpha^{(l)}_{i,j}$ if $j \in \mathcal{N}_{i}$ otherwise $\mathbold{A}^{(l)}_{i,j} =  0$, and $\sum_{i=1}^{N}\alpha^{(l)}_{i,j} = 1$. Then we have the following
proposition (See the proof in Appendix) that a single attention layer acts as a kind of random walking Laplacian smoothing.
\begin{prop}\label{prop:normLap}
	Let matrix ($\mathbold{I}  - \mathbold{A}$) be a random walk normalized Laplacian of the graph $\mathcal{G}$. And a single attention layer is equivalent to the Laplacian smoothing operation.
\end{prop}

Let $P$ be a transition probability matrix of a connected undirected graph $\mathcal{G}$ with $N$ nodes, $P^{(t)}(v_i, v_j)$ be the probability of being at node $v_j$ after $t$ steps walking in $\mathcal{G}$ if we start at $v_{i}$, and $d_{v}$ be the degree of node $v$. Then, we have the following theorem (See the proof in Appendix). 
\begin{theorem}\label{theo:smooththoery}
	If the graph $G$ has no bipartite components, there exists a random walk on $\mathcal{G}$ with transition probability matrix $P$, that converges to a unique stationary distribution $\pi$. That is, for any pair of nodes $\{v_i, v_j\}$, $\lim\limits_{t\rightarrow \infty} P^{(t)}(v_i, v_j) = \pi({v_j}) = \frac{d_{v_{j}}}{\sum_{k}^{N}d_{k}}$.
\end{theorem}
 We can view attention weight matrix $\mathbold{A}$ as a random walk transition probability matrix since $\mathbold{A}_{i,j}\geq 0$ and $\sum_{j=1}^{N} \alpha_{i,j} = 1$. Therefore, suppose there are $k$ connected components $\{C_{i}\}_{i=1}^{k}$ in the graph $\mathcal{G}$, according to Theorem \ref{theo:smooththoery}, by repeatedly applying random walking Laplacian smoothing multiple times (this is similar to increasing the depth of GAT), the features of the nodes in each connected component will converge to same values. Based on the cluster assumption in node classification that the nodes in same connected component tend to share same labels, the smoothing results in a easier classification problem. This is the reason why GAT works for node classification.
 
\xhdr{Different Attention Weights at Every Layer}
In practice, attention weight matrices vary in different layers. This is different from Theorem \ref{theo:smooththoery} which multiplies an identical matrix repeatedly. In fact, stacking multiple GAT layers together is equivalent to matrix-chain multiplication over multiple different attention weight matrices. We have Theorem \ref{theo:smooththoery2} (See the proof in Appendix) to demonstrate that GATs suffer from over-smoothing when they go deep, since that the attention matrix at each layer can be viewed as a transition probability matrix on the graph.
\begin{theorem}\label{theo:smooththoery2}
	Let $P^{(l)}$ ($l$$\geq$$1$) be a transition probability matrix of the connected undirected graph $\mathcal{G}$, corresponding to attention scores of $l$-th GAT layer, then $\lim\limits_{l\rightarrow \infty}\Pi_{i=1}^{l}P^{(l)} = \pi$, where $\pi$ is the unique stationary distribution in Theorem \ref{theo:smooththoery}.
\end{theorem}
In practice most graphs contain bridge nodes that connect different components with different labels. Theorem \ref{theo:smooththoery2} states that if we increase the depth of GAT, due to the boundary nodes, the aggregated node features of different components would become indistinguishable, leading to worse performance of deep GATs (See the observation of over-smoothing in Appendix). We call this phenomenon as \textbf{over-smoothing}.

\noindent\textbf{Multi-head Attention} is employed in GAT \cite{velivckovic2017graph}. Specially, $K$ independent attention heads are computed for feature aggregation at each layer, and the output of that layer is the concatenated outputs from all heads. To facilitate the analysis, we only focus on the attention aggregation and simplify Eq. \ref{eq:GAT} on each head as $\mathbold{X}^{(l,k)} = \mathbold{A}^{(l,k)}\mathbold{X}^{l-1}$, where $\mathbold{A}^{(l,k)}$ be the $k$-th (1$\leq$ $k\leq$ $K_{l}$) head attention matrix in $l$-th layer of GAT, $K_{l}$ is the head number of $l$-th layer. The output of $l$-th layer $\mathbold{X}^{l}$ = $\mathbin\Vert_{k=1}^{K_{l}}(\mathbold{A}^{(l,k)}\mathbold{X}^{l-1})$, where $\mathbin\Vert$ denotes concatenation along the column (hidden) dimension. 
By expanding this equation for the previous layer, we can get that each independent component $\mathbold{A}^{(l,i)}\mathbold{X}^{l-1}$ = $\mathbold{A}^{(l,i)}\mathbin\Vert_{k=1}^{K_{l-1}} (\mathbold{A}^{(l-1,k)}\mathbold{X}^{l-2}$) = $\mathbin\Vert_{k=1}^{K_{l-1}}(\mathbold{A}^{(l,i)}\mathbold{A}^{(l-1,k)}\mathbold{X}^{l-2})$. We can perform this expansion recursively for all layers. Therefore the output of $l$-th layer consists of multiple components, where each component can be viewed as a matrix-chain multiplication on $l$ attention matrices from different heads and layers. According to Theorem \ref{theo:smooththoery2}, these matrix-chain multiplications will converge to the unique distribution $\pi$ if $l \leftarrow \infty$. This means that multi-head attention GATs still suffer from over-smoothing problem if they go deep.

\begin{figure}[!ht]
	\centering
	\includegraphics[width=0.5\textwidth, height=4.5cm]{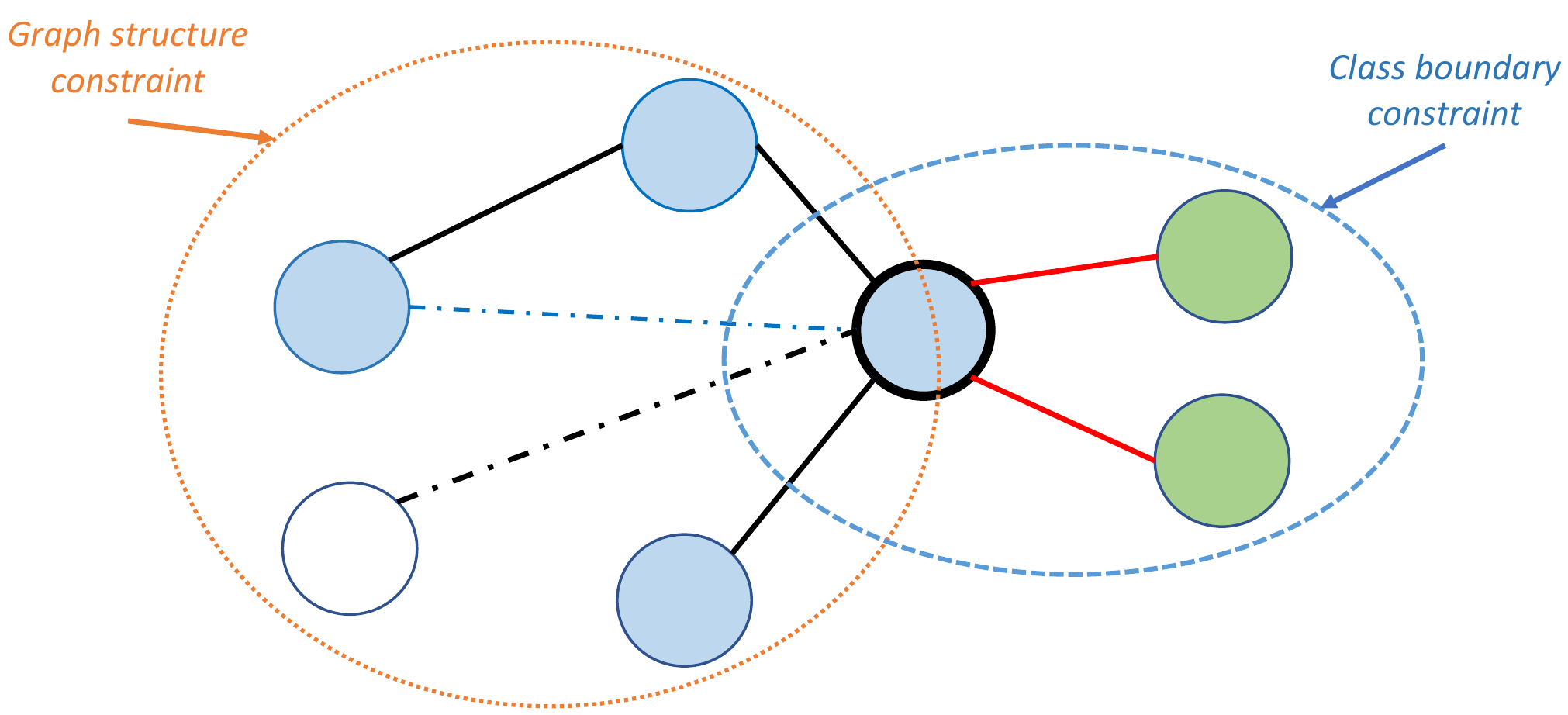}
	\caption{\textbf{Target node}: dark black solid bound circle; nodes with same colors share same class labels; white circle means any un-reachable node from the target node.
		\textbf{Left Orange Circle}: graph structure based constraint requiring there is a margin between attention from one-hop neighbors (black solid line) and that from multi-hop neighbors (blue dashed line) or unreachable nodes (black dashed line); \textbf{Right Blue Circle}: Class boundary based constraint requiring there is a margin between attention from neighbors with same class labels and the neighbors with different class labels.}
	\label{fig:constraints}
	\vspace{-1.5em}
\end{figure}

\xhdr{Residual connection} is an effective way to ensure good performance when increasing the depth of Convolutional Neural Networks \cite{he2016deep}. It has also been employed in GATs \cite{velivckovic2017graph,liu2018geniepath}. To formally prove the effect of residual connections, we introduce the concept of \textbf{Lazy Random Walk} as follows: 

Let $\mathbold{A}$ be a random walk transition probability matrix, we first define the transition probability matrix of a lazy random walk as $\mathbold{P} = \frac{\mathbold{I} + A}{2}$. At every step, the lazy random walk has $50\%$ probability of staying at the current node, and $50\%$ probability of moving away from it. Hence the residual connection $\mathbold{Y} = \mathbold{A}\mathbold{X} + \mathbold{X} = (\mathbold{I} + \mathbold{A})\mathbold{X}$ is a lazy random walk based smoothing up to a constant factor\footnote{The constant factor depends only on number of layers and is the same for all nodes}. Therefore if the lazy random walk $\mathbold{P}$ is viewed as a transition probability matrix, by Theorem \ref{theo:smooththoery2}, the features of all nodes in a connected component converge to the same values if more GAT layers are stacked. Moreover, the following Theorem in \cite{chung2005laplacians} answers how fast the  lazy random walk based smoothing process converges to a stationary distribution.

\begin{theorem}\label{theo:smoothingrate}
	Suppose that a strongly connected directed graph $\mathcal{G}$ on $n$ nodes has Laplacian eigenvalues $0=\lambda_{0}\leq \lambda_{1}\cdots \lambda_{n-1}$. Then $G$ has a lazy random walk with the rate of convergence of order $\frac{2}{\lambda_{1}}(-\log\min_{v}\pi(v))$. Namely, after at most $t\geq \frac{2}{\lambda_{1}}(-\log\min_{v}\pi(v) + 2c)$ steps, we have:
	\begin{equation*}
	\Delta(t) \triangleq \max_{1\leq i\leq n}(\sum_{1\leq j \leq n}\frac{(P^{(t)}(v_i, v_j) - \pi(v_j))^2}{\pi(v_j)})^{\frac{1}{2}} \leq e^{-c}.
	\end{equation*}
\end{theorem}

Theorem \ref{theo:smoothingrate} implies that it is difficult to prevent the over-smoothing of deep GAT by simply adding residual connections. This phenomenon has also been confirmed by experiments in \cite{liu2018geniepath}.

\section{Constrained Graph Attention Networks}
\label{sec:method}
To address the problems of overfitting and over-smoothing of GATs, we propose a framework called constrained graph attention networks (C-GATs) via adding constrains on both attention function and feature aggregate function. With these constraints, we can improve the generalization ability and alleviate the problem of over-smoothing of GAT. In the following, we first introduce two constraints on attention computation, which involves two margin based losses to guide the training of GNN. Then, based on the constrained attentions, we propose a new aggregation function, which chooses a subset of neighboring nodes based on attention weights rather than all neighbors for feature aggregation, to further reduce the over-smoothing of GAT.

\subsection{Margin based Constraint on Attention}
\label{subsec:marginConstraint}

To address the problem of over-fitting, we either make use of more data or use regularization techniques for attention function training. \cite{hamilton2017inductive} utilized graph structure data to guide the graph representation learning, which achieves impressive performance improvement for node classification. They required that the similarity between nearby nodes should be larger than those of disparate nodes. Nearby nodes are identified by a fixed-length random walk. This means that the graph structure is very important for graph representation learning. Inspired by this idea, the first constraint on attention function would be to induce the computed attention weights to reflect the graph structure.
More precisely, we require the attention weights between one-hop neighboring nodes be greater than those of disparate nodes (including multi-hop neighboring nodes). 
This can be viewed as a simplified version of~\cite{hamilton2017inductive}. 

We apply a second constraint on attention computation to address the over-smoothing in GAT. Over-smoothing occurs if a pair of nodes with different class labels are connected, as the information of different classes gets mixed via such pairs of nodes. To prevent the information communication, we require that the attention weights between nodes that shared the same class labels are greater than those weights between nodes that are with different class labels. This constraint is called the class boundary based constraint.

For a given node $v_i$, suppose $\mathcal{N}_{i}$ is the set of its one-hop neighboring nodes, and  $\mathcal{N}^{-}_{i}\subset \mathcal{N}_{i}$ and $\mathcal{N}^{+}_{i} \subseteq \mathcal{N}_{i}$ are the neighbors with different and same class labels to $v_i$ \footnote{$\subseteq$ is due to the self-loop connection.}. 
Fig. \ref{fig:constraints} gives an illustration of two margin-based constraints.

\begin{enumerate}[leftmargin=10pt]
	\item {\em Loss from Graph Structure based Constraint}:
	\begin{equation}\label{eq:structConstraint}
	\mathcal{L}_{g} = \sum_{i\in \mathcal{V}} \sum_{j\in \mathcal{N}_{i}\setminus \mathcal{N}^{-}_{i}}\sum_{k\in (\mathcal{V}\setminus\mathcal{N}_{i})}\max(0, \phi(v_i, v_k) + \zeta_{g} - \phi(v_i, v_j))
	\end{equation}
	
	\item {\em Loss from Class Boundary Constraint}:
	\begin{equation}\label{eq:boundConstraint}
\mathcal{L}_{b} = \sum_{i\in \mathcal{V}} \sum_{j\in \mathcal{N}^{+}_{i}}\sum_{k\in \mathcal{N}^{-}_{i}}\max(0, \phi(v_i, v_k) + \zeta_{b} - \phi(v_i, v_j))
	\end{equation}
\end{enumerate}
where $\zeta_{g}$$\geq$0 and $\zeta_{b}$$\geq$0  are slack variables which control the margin between attention values. $\phi(v_i, v_j)$ is then attention function. Let $\mathbold{h}_i$$\in$$\mathbb{R}^{f}$ and $\mathbold{h}_j$$\in$$\mathbb{R}^{f}$ be the features of nodes $v_i$ and $v_j$, we use  $\phi(\mathbold{h}_i, \mathbold{h}_j) = \text{LeakyReLU}(\text{MLP}(\mathbold{W}_{r}(\mathbold{h}_i || \mathbold{h}_j))$ to compute the attention between two nodes $v_i$ and $v_j$, $\mathbold{W}_r \in \mathbb{R}^{f'\times 2f}$ is a trainable matrix.

\xhdr{Adaptive Negative Sampling for GNN Training} Negative sampling has been proved to be an effective way to optimize the loss function $\mathcal{L}_{g}$ in Eq. \ref{eq:structConstraint}. A uniform sampling of negative examples would suffer the problem of inefficiency since many negative samples are easy to classify as the model training goes on. And these negative example would not provide any meaningful information to the model training. Here we propose a new approach to choose negative examples adaptively for each layer. 

For a given node, we assume that the important negative sample nodes are the non-neighboring nodes which have large contribution in feature aggregation to the other nodes. This means that the more contribution of a node for feature aggregation, the more possible it is a good negative candidate node. Therefore, we apply importance sampling to choose negative sample nodes. The importance of a node can be estimated by Proposition \ref{prop:negsample} (See the proof in Appendix).

\begin{prop}\label{prop:negsample}
	The importance of a node $v_i$ to feature aggregation of $\mathcal{G}$ in $l$-th layer is proportional to 
	$\sum_{j=1}^{N}{\alpha}^{(l)}_{j,i}$.
\end{prop}

According to Proposition \ref{prop:negsample}, we construct a negative sampling as follows: we use a  weighted random sampler. Weights can be efficiently computed based on the attention matrix by summing the attention weights of one column in $\mathbold{A}$.

With these two constraints, we can optimize the following loss functions for node classification:
	\begin{equation}\label{eq:constraintLoss}
		\mathcal{L} = \mathcal{L}_{c} + \lambda_{g} \mathcal{L}_{g} + \lambda_{b} \mathcal{L}_{b},
	\end{equation}
where $\mathcal{L}_{c}$ represents the loss derived from the node classification error (e.g., cross entropy loss for multi-class node classification) and $ \lambda_{g}\geq 0$ and $ \lambda_{b}\geq 0$ are two weight factors to make trade-offs among these losses, which are data dependent.

\subsection{Constrained Feature Aggregation}\label{subsec:feaAgg}

According to the analysis of GATs, the over-smoothing of GAT occurs from the information mixing along the bridging nodes connecting two different clusters. In this section, we propose a constrained feature aggregate function to prevent such information mixing. For each node, the aggregate function only makes use of the features from the neighbors with top $k$ attention weights rather than all neighbors.  
From the constraint on attention computation in Eq. \ref{eq:boundConstraint}, 
the attention weights of the nodes from different classes should be small. Therefore, picking up nodes with top $k$ attention weights would not only keep the smoothing effect of features of the nodes within same class but also drop edges that connect different classes due to small attention weights.

Note that the parameter $k$ makes a trade-off between smoothing and over-smoothing. In principle, for a connected graph, if we guarantee that the all the selected top $k$ nodes could still form a connected graph, then we can keep a smoothing effect of GNN models. The top $k$ based feature aggregator can be viewed as a sub-graph selector, which selects different sub-graphs for Laplacian smoothing in different layers. Therefore, it alleviates the over-smoothing in existing GAT models which always use the same graph structures for feature aggregation. The top-$k$ selection based feature aggregate function helps the model go deeper with more layers. 
However, a small $k$ would allows GAT go deeper, but might cause high variance in aggregation. Therefore, in practice, we should select the parameter $k$ to make a trade-off between these two aspects. The results of sensitive analysis of $k$ in experimental section demonstrate such trade-off.

\noindent\textbf{Comparison to Attention Dropout in GAT.} Attention dropout randomly selects a proportion of attentions for feature aggregation \cite{velivckovic2017graph}. Experiments have  confirmed that it is helpful for GAT training for small graphs. It can also be viewed as a process of selecting different sub-graphs for laplacian smoothing. This motivation is similar to our constrained feature aggregation. However, with the random dropout mechanism, it is still difficult to prevent the information mixing along the bridging nodes, since the bridges will only be removed with probably equal to the dropout probability, leading to poor performance. As an example, see Figure \ref{fig:results_comparison} (a), 
which shows that the performance of GAT with dropout is significantly lower when we add noisy edges in the graphs. The noisy edges here act as bridges between nodes with different label classes. We observe that GAT with dropout cannot effectively cope with the more pronounced oversmoothing due to noisy edges. In contrast, our method still performs  well on these noisy datasets.

\section{Experiments}\label{sec:experiments}
\xhdr{Data Set} We evaluate the performance of the proposed algorithm C-GAT (\underline{C}onstrained \underline{G}raph \underline{A}ttention ne\underline{T}works) on four node classification benchmarks: (1) categorizing academic papers in the citation network datasets: Cora, Citeseer and Pubmed \cite{sen2008collective}; (2) classifying protein functions across various biological protein-protein interaction (PPI) graph \cite{zitnik2017predicting}. Table \ref{table:dataset} summarizes the statistical information of these datasets. Our experiments are conducted over standard data splits \cite{huang2018adaptive}. Following the supervised learning scenario, we use all the labels in the training examples for model training.
\begin{table*}[ht!]
	\centering
	\caption{Statistical Information on Benchmarks}
	\begin{threeparttable}
		\begin{tabularx}{0.65\textwidth}{r c c c c c} 
			\hline
			Name & Nodes & Edges & Classes & Node features   & Train/Dev/Test\\
			\hline
			Cora\tnote{a} & 2708 & 5429 & 7 & 1433  & 1,208/500/1,000\\
			\hline
			Citeseer\tnote{a} & 3327 & 4732 & 6& 3703 & 1,827/500/1000\\
			\hline
			Pubmed\tnote{a} & 19717 & 88651 & 3 & 500 &  18,217/500/1,000\\
			\hline
			PPI\tnote{b} & 56944\tnote{$\ast$} & 818716 & 121\tnote{$\star$}& 50 & 20/2/2\tnote{$\diamond$}\\
			\hline
		\end{tabularx}
		\begin{tablenotes}
			 \item a: transductive problem;  b: inductive problem; $\star$: multi-label; $\ast$: total nodes in 24 graphs; $\diamond$: 20 graphs for train, 2 graphs for validation and 2 graphs for test.
		\end{tablenotes}
	\end{threeparttable}
	\label{table:dataset}
	\vspace{-1.5em}
\end{table*}

\xhdr{Hyper-parameter Settings} For three transductive learning problems, we use two hidden layers with hidden dimension as 32 for Cora, 64 for Citeseer, and three hidden layers with hidden dimension 32 for Pubmed; we set the number of neighbors $k$ used in feature aggregate function as 4 for Cora, Citerseer, and 8 for Pubmed. For the inductive learning problem PPI, we use three hidden layers with hidden dimension 128, and set $k$ as 8.  
We make use of Adam as the optimizer and perform hyper-parameter search for all baselines and our method over the same validation set. The set of margin values ($\zeta_{g}, \zeta_{b}$) used in ($\mathcal{L}_{g}$, $\mathcal{L}_{b}$) is \{0.1, 0.2, 0.3, 0.5\}\, and the trade-off factor ($\lambda_{g},
\lambda_{b}$) of two losses is set as \{1, 2, 5, 10\}, learning rate is set as \{0.001, 0.003, 0.005, 0.01\} and $\ell_{2}$ regularization factor is set as \{0.0001, 0.0005, 0.001\}. We train all models using early stopping with a window size of 100.



\xhdr{Baselines} We compare our C-GAT with the following representative GNN models: GCN~\cite{kipf2016semi}, GraphSAGE~\cite{hamilton2017inductive}, and Graph Attention Network (GAT)~\cite{velivckovic2017graph}. For transductive learning problems, since the results in~\cite{kipf2016semi,hamilton2017inductive,velivckovic2017graph} were from semi-supervised data setting, we present results of node classification based on our experiments following the same hype-parameter settings reported in these papers. We take the best GraphSAGE results from different pooling strategies~\cite{huang2018adaptive,velivckovic2017graph}. Meanwhile, for inductive learning problem PPI, we also compare C-GAT with other two representative attention based GNN models GaAN~\cite{zhang2018gaan} and GeinePath~\cite{liu2018geniepath}.

\xhdr{Evaluation Settings} We use the same metrics in GAT \cite{velivckovic2017graph} for classification performance evaluation. Specially, classification accuracy is collected over Cora, Citeseer and Pubmed, and Micro $F_{1}$ is collected over the multi-label classification problem PPI. We report the mean and standard deviation of these metrics collected for 10 runs of the model under different random seeds.

\subsection{Experimental Results}
We investigate the proposed algorithm C-GAT in the following four aspects: (1) classification performance comparison; (2) robustness which indicates whether the C-GAT is able to overcome the overfitting problem, and improve generalization on unseen graph structure; (3) depth of GNN models to demonstrate whether C-GAT can prevent the over-smoothing problem suffered by GAT and (4) sensitive analysis of the number of neighbors $k$ used in feature aggregate functions. 
\begin{table*}[h!]
	\centering
	\normalsize
	\caption{Classification Accuracy Ablation and Comparison}
	\begin{threeparttable}
		\begin{tabularx}{0.65\textwidth}{r |  r c c c c} 
			\hline
			& Methods & Cora & Citeseer & Pubmed & PPI\tnote{$\ast$}\\
			\hline
		\multirow{4}{*}{\rotatebox{90}{\hspace*{-2pt} GNN}} & GCN & $86.3 \pm 0.4$ & $75.6 \pm 0.3$ & $86.8 \pm 0.3$ & 71.0\tnote{$\dagger$}\\
			& GraphSAGE & $86.9 \pm 0.4$ & $76.5\pm 0.4$ & $85.7 \pm 0.4$ & 76.8\tnote{$\ddagger$} \\
			& GAT & $87.2 \pm 0.3$ &  $77.3\pm 0.3$ & $87.0\pm 0.3$  & $97.3 \pm 0.02$ \\
			\cline{2-6}
			& C-GAT  & $\mathbf{88.4} \pm 0.3$ & $\mathbf{79.9} \pm 0.3$ & $\mathbf{87.6} \pm 0.3$ & $\mathbf{98.8} \pm 0.05$\\
			\hline
			\multirow{4}{*}{\rotatebox{90}{\hspace*{-2pt} Ablation}} & w/o $\mathcal{L}_{g}$ & $88.3 \pm 0.2$ & $78.7 \pm 0.2$ &  $87.2 \pm 0.3$ & $98.1 \pm 0.04$\\
			& w/o $\mathcal{L}_{b}$ & $88.2  \pm 0.3$ & $79.3  \pm 0.3$ & $87.2  \pm 0.2$ & $97.9  \pm 0.04$\\
			& w/o top $k$ & $88.4  \pm 0.2$& $78.5  \pm 0.2$ & $87.3  \pm 0.2$ & $97.5  \pm 0.03$\\
			& w/o NINS\tnote{$\star$}  & $88.2 \pm 0.3$ & $78.9  \pm 0.2$  & $87.4  \pm 0.3$ & $98.2  \pm 0.04$ \\
			\hline
		\end{tabularx}
		\begin{tablenotes}
			\item [$\ast$] The accuracy of the attention based GNN models on \textbf{PPI: GaAN}~\cite{zhang2018gaan} \textbf{98.7 $\pm$ 0.02} and \textbf{GeinePath}~\cite{liu2018geniepath} \textbf{97.9}, respectively;
			\item [$\dagger$] The best accuracy of GCN on PPI reported in \cite{liu2018geniepath};
			\item [$\ddagger$] The best accuracy of GraphSAGE on PPI reported in \cite{velivckovic2017graph};
			\item [$\star$] NINS: Node Importance based Negative Sampling.
		\end{tablenotes}
	\end{threeparttable}
	\label{table:classificationResults}
	\vspace{-0.5em}
\end{table*}

\xhdr{\textbf{Classification}} We report results of performance comparison and ablation study in Table \ref{table:classificationResults}. From this table, we observe that:

\noindent(a) Our model C-GAT performs consistently better than all baseline models GCN, GraphSAGE and GAT across all benchmarks. Specifically, we improve upon GAT with absolute accuracy gain of
1.2\%, 2.6\%, 0.6\% and 1.5\% on Cora, Citeseer, PubMed and PPI, respectively. Especially for the inductive learning problem PPI, we get the new state-of-the-art classification performance~\cite{zhang2018gaan,liu2018geniepath}.

\noindent(b) Using ablation studies in Table \ref{table:classificationResults}, we observe that the proposed constraints and $k$ selected neighborhood based aggregation function achieves especially large gain on PPI. In inductive learning setting such as PPI, the testing graph is completely un-seen, where overfitting of attention is especially significant. This suggests that the proposed constraints make the attention functions can generalize well to un-seen graph structure.
The last row in Table \ref{table:classificationResults} shows the classification accuracy of C-GAT with uniform negative sampling instead of the adaptive node-importance based negative sampling method proposed in this paper. 
The results imply that involving the nodes' importance into negative sampling brings benefit for the training of C-GAT.

\begin{figure*}[!h]
	\includegraphics[width=1.0\textwidth]{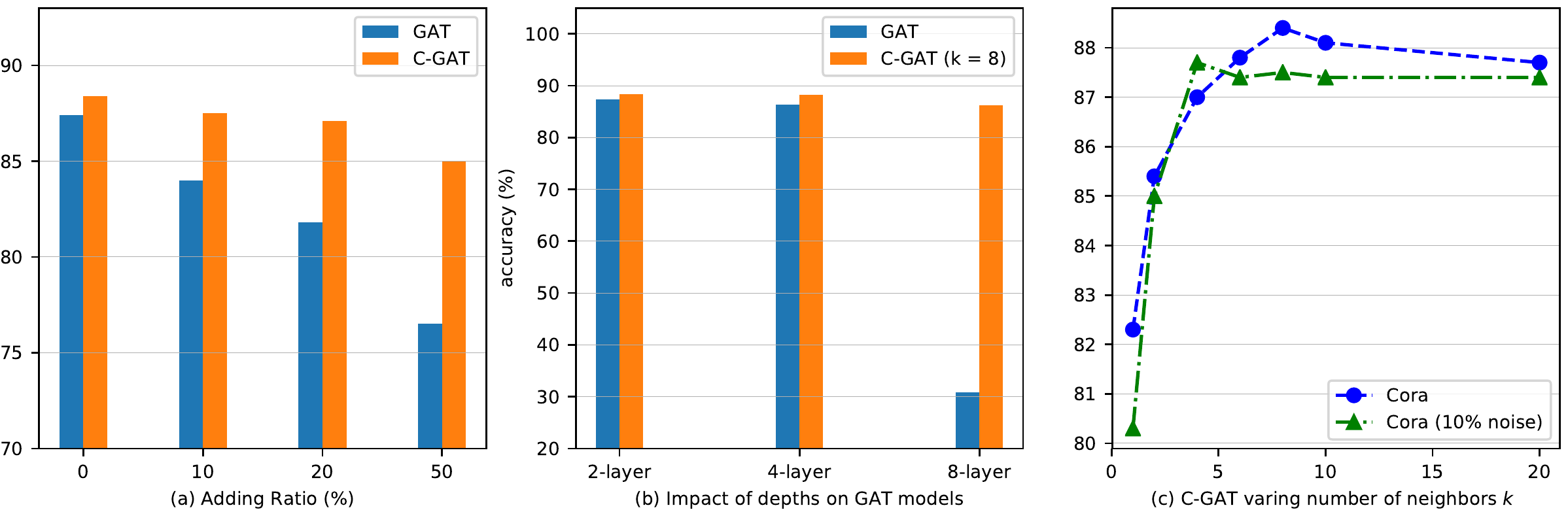}
	\caption{Experiments on Cora. \textbf{Left: Robustness Analysis}:  Train on original graphs and perform testing on graphs by adding edges randomly. For adding edges, we first randomly select a set of nodes according to a given sampling ratio, and then random add one edge on these nodes. \textbf{Middle: Deeper GAT}: Classification performance comparison between GAT and C-GAT with different depth; \textbf{Right: Sensitive Analysis} Impact of number of neighbors $k$ on classification performance of C-GAT.}
	\label{fig:results_comparison}
	\vspace{-1.5em}
\end{figure*}

\xhdr{\textbf{Robustness Analysis}} To demonstrate the robustness of C-GAT, \emph{i.e.} whether the induced attention function is robust to the graph structure, we conduct experiments by perturbing edges in ``Cora'' test data. Fig. \ref{fig:results_comparison} (a) presents the experimental results. We observe that:


By random adding edges in testing stage, GAT shows a significant descending trend when increasing the ratio of adding edge. Randomly adding edges might connect different classes together. This aggravates the over-smoothing problem of GAT. However, our algorithm C-GAT still get good predictions even when the ratio of adding edges is up to 50\%.  Because of the class boundary constraint, C-GAT would assign small attention values on these boundary edges. Moreover, the proposed $k$ selected neighbor based feature aggregation function would further eliminate such negative impacts. These results demonstrate better generalization of our C-GAT than GAT on unseen graphs (see more results on robustness in Appendix).


\xhdr{\textbf{Deeper GAT}} Fig. \ref{fig:results_comparison} (b) compares C-GAT and GAT with different depths on ``Cora''. In contrast to the degradation of GAT with deeper layers due to more significant oversmoothing, Our proposed C-GAT maintains good classification performances with increasing attention layers. Again these results show that C-GAT is able to effectively overcome the problem of oversmoothing. This allows the applications of C-GAT in graph-level tasks where depth is critical~\cite{bunz2017graph}.

\xhdr{\textbf{Sensitive Analysis of Neighbor Number $k$}} We also analyze the sensitivity of the hyper-parameter $k$, which controls the aggregation step based on high attention weights. We conduct experiments of C-GAT by varying $k$ in range \{1, 2, 4, 6, 8, 10, 20\} on ``Cora''.  
We randomly add 10\% edges to ``Cora'' to increase the chance of information propagation among different classes and then investigate how to set $k$ on these noisy graphs. The right sub-fig in Fig. \ref{fig:results_comparison} (c) gives the impact of $k$ on classification accuracy of C-GAT on ``Cora'' and ``Cora with 10\% noisy edges''. 
 
We observe that the classification accuracy first increases to a peak value and stabilizes or slightly decreases. 
This means that $k$ plays a role of making trade-off between under-smoothing (not enough smoothing to tackle noise) and over-smoothing. For example, in Fig.~\ref{fig:results_comparison} a smaller $k = 4$ would be best for noisy graphs, whereas a larger $k = 8$ achieves the best performance on original graphs.


\section{Conclusion}
In this paper we provide analysis of the weakness of GAT models: over-fitting of attention function and over-smoothing of node representation on deeper model.
We propose a novel approach called constrained graph attention Network (C-GAT), to address the overfitting and over-smoothing issues of GAT by guiding the attention during GAT training using margin-based constraints. In addition, a layer-wise adaptive node-and-edge sampling approach is proposed for augmenting attention training with effective negative examples. Furthermore, to alleviate the over-smoothing problem we propose a new feature aggregate function which only selects the neighbors with top K attention weights rather than all the neighbors. Extensive experiments on common benchmark datasets have verified the effectiveness of our approach, and show significant gains in accuracy on standard node classification benchmarks, especially on deeper models and noisy tests, compared to the state-of-the-art GAT models. A particularly interesting direction for future work is to explore more effective constraints in attention computation of GAT for other downstream tasks (e.g., link prediction).


\appendix






\section{Proof of Proposition \ref{prop:normLap}}\label{app:singleLayer}

Before we give the proof, we first introduce the concepts of random walking normalized Laplacian and Laplacian smoothing as follows.

\textbf{Random walking Normalized Laplacian} Let $\mathbold{A}_{N\times N}$ be the attention weight matrix, $\mathbold{D} = \text{diag}(d_1, d_2, \cdots, d_{N})$ and $d_i = \sum_{j=1}^{N}\mathbold{A}_{i,j}$, then the graph Laplacian of $\mathcal{G}$ is defined as $\mathbold{L} = \mathbold{D} - \mathbold{A}$. And $\mathbold{L}_{rw} = \mathbold{D}^{-1}\mathbold{L}$ is the random walking normalized Laplacian of $\mathcal{G}$.

\textbf{Laplacian Smoothing} \cite{li2018deeper} on each row of the input feature matrix $\mathbold{X}$ is defined as:
\begin{equation}\label{eq:lopsmooth}
\mathbold{y}_{i} = (1 - \lambda)\mathbold{x}_i + \lambda \sum_{j}^{N}\frac{\alpha_{i,j}}{d_{i}}\mathbold{x}_{j},
\end{equation}
where $0 < \lambda \leq 1$ is a parameter to controls the smoothness, \emph{i.e.} the importance weight of the node's features with respect to the features of its neighbors. We can rewrite the Laplacian smoothing in Eq. \ref{eq:lopsmooth} in matrix form: 
\begin{equation}
\mathbold{Y} = (\mathbold{I} - \lambda \mathbold{D}^{-1} \mathbold{L}) \mathbold{X}  = (\mathbold{I} - \lambda \mathbold{L}_{rw}) \mathbold{X}
\end{equation}

\begin{proof}
	As $\mathbold{A}_{N\times N}$ is the attention weight matrix, $d_i = \sum_{j}^{N}{\alpha_{i,j}} = 1$, then we can get that $\mathbold{D} = \mathbold{I}$. The random walk normalization of $\mathcal{G}$ is $\mathbold{L}_{rw} = \mathbold{D}^{-1}\mathbold{L}$ = $\mathbold{I}^{-1}(\mathbold{I} -  \mathbold{A})$ = $\mathbold{I} -  \mathbold{A}$.
	
	We can rewrite the graph attention operation $\mathbold{Y} = \mathbold{A}  \mathbold{X}$ as $\mathbold{Y} = (\mathbold{I}  - \mathbold{L}_{rw}) \mathbold{X}$. According to the formulation of Laplacian smoothing in Eq. \ref{eq:lopsmooth}, we can conclude that graph attention is a special form of Laplacian smoothing with $\lambda = 1$. 
\end{proof}


\section{Proof of Theorem \ref{theo:smooththoery}}\label{append:proofOfTheory1}
\begin{proof}
	(1) We can view the random walk on graph $G$ as a Markov chain with $P$.  As $G$ is undirected, connected and non-bipartite graph, the Markov chain is ergodic \cite{randall2006rapidly,lovasz1993random}. And any finite ergodic Markov chain converges to a unique stationary distribution $\pi$ \cite{randall2006rapidly}.
	(2) According to Perron-Frobenius Theorem \cite{horn2012matrix,chung2005laplacians}, such stationary distribution is just the Perron vector of $P$. And for the undirected graph, its Perron vector w.r.t. $v_i$ is $d_{v_i}/\sum_{j}^{N}d_{v_j}$.
\end{proof}

\section{Proof of Theorem \ref{theo:smooththoery2}}\label{append:proofOfTheory2}
\begin{proof}
	 (1) Let $P^{(l)}$ be the transition matrix over the graph $\mathcal{G}$, corresponding to attention weight matrix of $l$-th layer of GAT. According to Theorem 1, the random walk on the graph $\mathcal{G}$ with $P^{(l)}$ converges to a unique stationary distribution which depends on the degrees of the graph regardless of $P^{(l)}$. i.e., $\pi_{l} = \pi$ where $\pi_{l}$ denotes the stationary distribution w.r.t. $P^{(l)}$ and $\pi$ is the unique stationary distribution. 
	 (2) Let $f^{k}_{i}$ be the $i$-th row of $\Pi_{t=1}^{k}P^{(k)}$, according to the converge analysis of random walk in \cite{randall2006rapidly}, we have $||f^{k}_{i} - \pi_{k}||$ $\leq$ $\lambda_{k} ||f^{k-1}_{i} - \pi_{k}|| = \lambda_{k} ||f^{k-1}_{i} - \pi_{k_1}||$ as $\pi_{k} = \pi_{k-1}$, where $\lambda_{k}$ is the mixing rate of random walk with $A_{k}$. By exploring the equation recursively, $||f^{k}_{i} - \pi||$ $\leq$ $\lambda_{k} ||f^{k-1}_{i} - \pi||$ $\leq\cdots \leq \Pi_{t=1}^{k}\lambda_{t}||f^{1}_{i} - \pi||$. Moreover, for strongly connected graph, the mixing rate $\lambda_{t} \in (0, 1)$ according to \cite{randall2006rapidly}. Then, $\lim\limits_{k\rightarrow \infty}||f^{k}_{i}-\pi|| = 0$. i.e., $\lim\limits_{k\rightarrow \infty}f^{k}_{i} = \pi$. 
\end{proof}

\section{Observation of Over-Smoothing on Data ``Citeseer''}\label{append:oversmooth}

\begin{figure*}[!h]
	\includegraphics[width=1.0\textwidth]{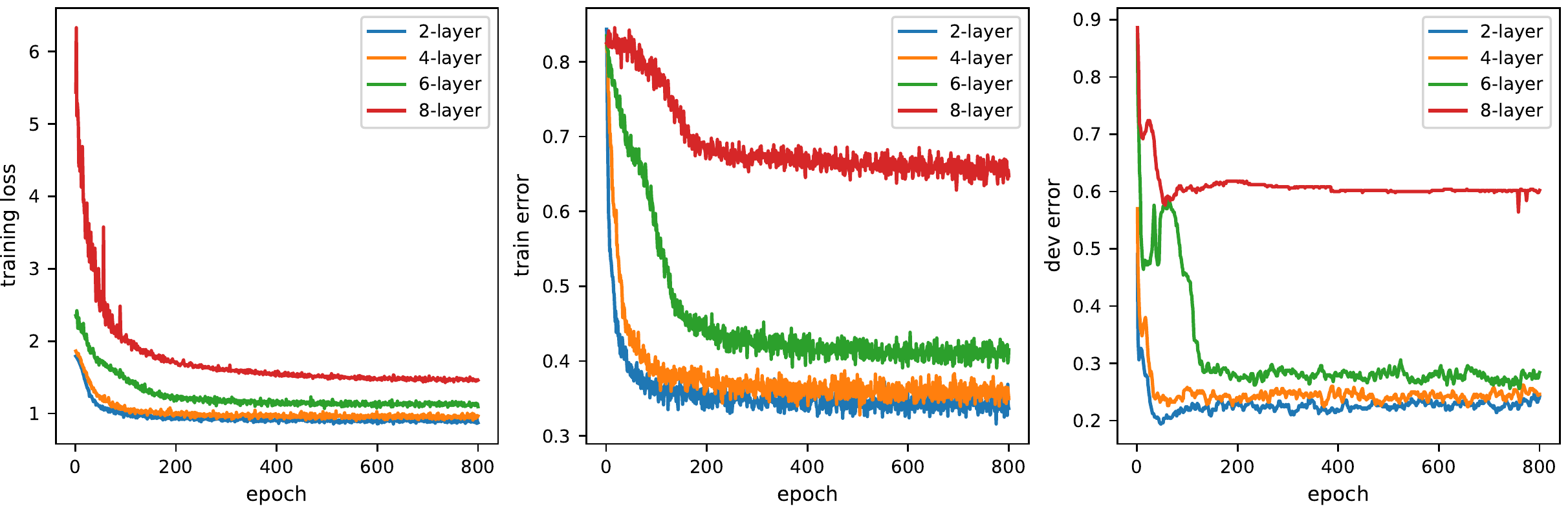}
	\caption{Training loss (left), training error (middle) and validation error (right) on Citeseer with 2-layer, 4-layer, 6-layer and 8-layer GAT models. The deeper network has higher training error, and thus validation error.}
	\label{fig:oversmoothCiteseer}
\end{figure*}

Fig. \ref{fig:oversmoothCiteseer} shows the training loss, training error and the validation error of GAT models with different layers on  benchmark dataset ``Citeseer'' (See detailed information of the data in Table \ref{table:dataset}).  From this figure, we can observe that the deeper networks can still converge, but a performance degradation problem occurs: with the depth increasing, the accuracy degrades. In this paper, we demonstrate that such performance degradation is mainly due to over-smoothing effect of deeper GAT models.

\begin{figure*}[!h]
	\includegraphics[width=1.0\textwidth]{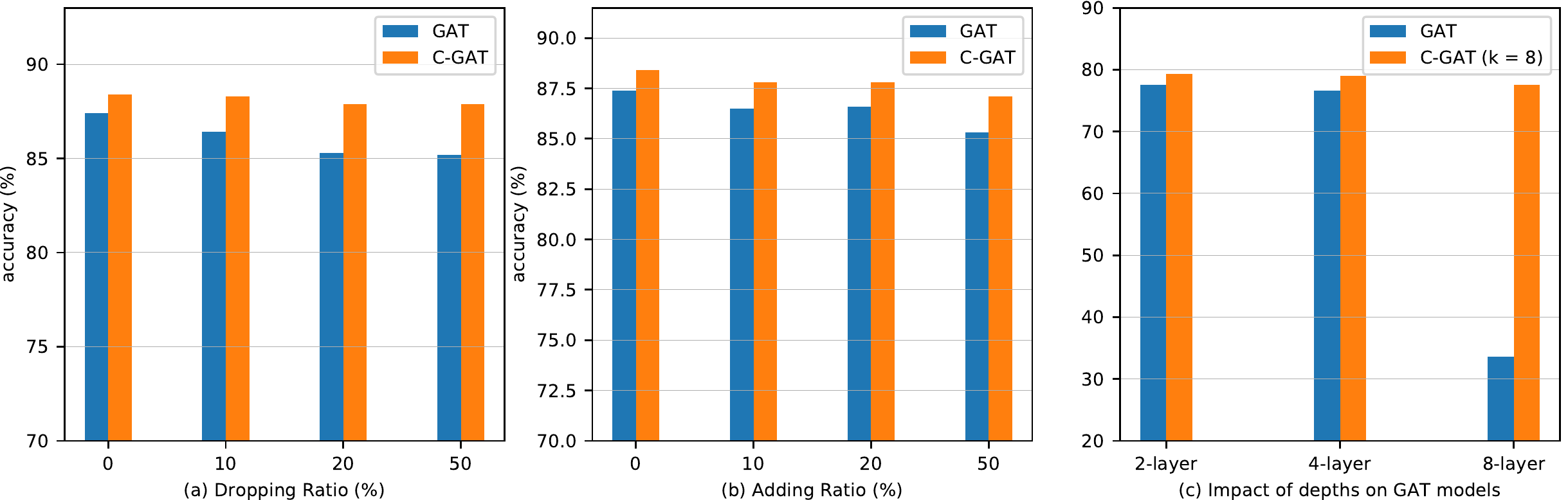}
	\caption{\textbf{Left}: randomly dropping edges in training stage and performing test on the original graph over ``Cora''; \textbf{Middle}: Randomly adding edges in training stage and performing test on the original graph ``Cora''. For adding edges, we first randomly select a set of nodes according to a given sampling ratio, and then random add one edge on these nodes. \textbf{Right}: Classification performance comparison between GAT and C-GAT with different depth on ``Citeseer''}
	\label{fig:appendAnalysis}
\end{figure*}

\section{Proof of Proposition \ref{prop:negsample}}\label{append:importantSammpling}

\begin{proof}
	Let's first review the feature aggregate function in GAT:.
	\begin{equation}
	\mathbold{h}_{i}^{(l+1)} = \delta(\sum_{j\in \mathcal{N}_{i}}\alpha^{(l)}_{i,j}\mathbold{W}^{(l)}\mathbold{h}_{j}^{(l)}) =  \delta(\sum_{j=1}^{N}\hat{\alpha}^{(l)}_{i,j}\mathbold{W}^{(l)}\mathbold{h}_{j}^{(l)}), 
	\end{equation}
	where $\hat{\alpha_{i,j}} = \alpha_{i,j}$ if $j \in \mathcal{N}_i$, otherwise $\hat{\alpha_{i,j}} = 0$. We can view $\hat{\alpha_{i,j}}$ as the importance of $v_j$ of $v_i$ given the graph with features $\mathbold{H}^{(l)} = [\mathbold{h}_1, \mathbold{h}_2, \cdots, \mathbold{h}_{N}]^{T}$. We can rewrite it as a form of conditional probability $\hat{\alpha}_{i,j} = p(v_{j}|v_{i}, \mathcal{G}, \mathbold{H}^{(l)})$. If we define $q(v_{i}|v_1, v_2, \cdots, v_N)$ (denoted as $q(v_i)$ for simplification) as the probability of sampling $v_i$ given all the nodes of the current layer, then we get $\hat{\alpha}^{(l)}_{i,j} = \frac{p(v_j|v_i)}{q(v_i)}$.
	Then, according to Bayes's formula, we can get $q(v_{i}|v_1, v_2, \cdots, v_N) \propto \sum_{j=1}^{N}{\hat{\alpha}}^{(l)}_{j,i} = \sum_{j=1}^{N}{\alpha}^{(l)}_{j,i}$.
	\end{proof}

\section{Experimental Results of Robust Analysis and Deeper GAT}\label{append:experimentresults}
To evaluate the robustness of C-GAT, in particular, whether the induced attention function is robust to the graph structure, we conduct experiments by perturbing edges in ``Cora'' data. Fig. \ref{fig:appendAnalysis} presents the experimental results. From this figure, we can observe that:


(a) By randomly dropping some edges in training stage (see Fig. \ref{fig:appendAnalysis} (a)), C-GAT get a relative stable performance when increasing the ratio of dropped edge. In contrast, the performance of GAT shows a descending trend. This is because of that, for a missing edge in testing stage, the attention value w.r.t. this edge in C-GAT is still convincible as the two constraints. That is, if the missing edge connected two nodes share same labels, according to the constraints, the attention weight will be higher and results in a better smoothing operator. In contrast, if the missing edge connected two nodes with different labels, because of proposed constraints and proposed feature aggregation function, the impact of such edge can be eliminated as well. In contrast, for GAT without these constraints, there is still information propagation no matter the missing edge lies in classification boundary or not, and even assign large attention values for the classification boundary edges, and lead to over-smoothing.

(b) By random adding some edges in training stages (see Fig. \ref{fig:appendAnalysis} (b)),the performance of C-GAT still keeps relative stable but GAT's performance decreases when increasing the ratio of adding edges. This is because of that, the randomly adding edges might connect different classes together. This will result in more information propagation among different classes and easily lead to the over-smoothing. This hurts the quality of the training data. The constraints in C-GAT can be viewed as a data cleaner which can improve the quality of the training data. In contrast, GAT has no such ability and leads to the induced model perform worse in testing stage.

(c) Compares C-GAT and GAT with different depths on ``Citeseer''. Our proposed C-GATs maintain good classification performances with increasing attention layers. Again these results prove over-smoothing is not an issue for C-GAT.
\end{document}